\pdfoutput=1

\documentclass[11pt]{article}

\usepackage{ACL2023}

\usepackage{times}
\usepackage{latexsym}

\usepackage[T1]{fontenc}

\usepackage[utf8]{inputenc}

\usepackage{microtype}

\usepackage{inconsolata}

\usepackage{amsmath,amssymb,amsthm}
\usepackage{xytree}
\usepackage{xcolor}
\usepackage{graphicx}
\usepackage[all]{xypic}
\usepackage{bm}
\usepackage{parskip} 
\usepackage{enumitem} 
\usepackage{pdfpages}
\usepackage{comment}
\usepackage{nameref}
\usepackage{pdfcomment}
\usepackage{titlesec}
\usepackage{tikz}
\usepackage{tikz-cd}
\usepackage{ocgx2}
\usepackage{hyperref}
\hypersetup{
	colorlinks=true,   
	linkcolor=blue,    
	filecolor=magenta, 
	urlcolor=cyan,     
	citecolor=green    
}
\usepackage{graphicx}
\usepackage{booktabs}
\usepackage{array}

\theoremstyle{definition}

\theoremstyle{remark}

\title{Adapting Feature Attenuation to NLP}

\author{Tianshuo Yang \\ University of Michigan \\
  \texttt{yangts@umich.edu} \\\And 
  Ryan Rabinowitz \\
  University of Colorado \\
  Colorado Springs \\
  \texttt{rrabinow@uccs.edu} \\\And 
  Terrance E. Boult, Jugal Kalita \\ 
  University of Colorado \\
  Colorado Springs \\
  \texttt{tboult@vast.uccs.edu,} \\
  \texttt{jkalita@uccs.edu} }

\begin{document}
\maketitle

\begin{abstract}
Transformer classifiers such as BERT deliver impressive closed‑set accuracy, yet they remain brittle when confronted with inputs from unseen categories—a common scenario for deployed NLP systems. We investigate Open‑Set Recognition (OSR) for text by porting the \emph{feature attenuation hypothesis} from computer vision to transformers and by benchmarking it against state-of-the-art baselines. Concretely, we adapt the COSTARR framework---originally designed for classification in computer vision---to two modest language models (BERT (base) and GPT-2) trained to label 176 arXiv subject areas. Alongside COSTARR, we evaluate Maximum Softmax Probability (MSP), MaxLogit, and the temperature‑scaled free‑energy score under the OOSA and AUOSCR metrics.  Our results show (i) COSTARR extends to NLP without retraining but yields no statistically significant gain over MaxLogit or MSP, and (ii) free-energy lags behind all other scores in this high‑class‑count setting.  The study highlights both the promise and the current limitations of transplanting vision‑centric OSR ideas to language models, and points toward the need for larger backbones and task‑tailored attenuation strategies.
\end{abstract}

\section{Introduction}

Transformer models, exemplified by BERT and GPT-2, are widely used for text classification and many other NLP tasks~\cite{ComparingBERT}.  Most of these models are trained under a \emph{closed‑set} assumption: every test document is expected to fall into one of the categories seen during training.  In practical settings this assumption often breaks.  Search engines must tag papers from brand‑new research areas, legal‑tech systems face unfamiliar case types, and content filters encounter fresh slang or memes.  When a model meets such unseen classes, it can produce confident yet incorrect predictions, which is especially risky in sensitive domains like healthcare or law~\cite{MovingTowardsOpenSetIncrementalLearning,Awholisticviewofcontinuallearning...}.  Open‑Set Recognition (OSR) seeks to address this issue by asking a classifier to (i) label known inputs and (ii) flag inputs that come from unknown classes~\cite{TowardsOpenSetDeepNetworks}.  Two obstacles make OSR difficult for transformers: soft‑max scores tend toward over‑confidence, and the final linear head may suppress latent dimensions that still hold useful signals for novelty detection.

In computer vision, the \emph{feature attenuation hypothesis} suggests that combining features \emph{before} and \emph{after} the classification layer can improve detection of unknowns.  COSTARR implements this idea and has produced strong open‑set results for image models~\cite{COSTARR}.  Whether the same concept helps in NLP is an open question.  In this work we adapt COSTARR to two medium‑size language models---BERT (base) and GPT-2---fine‑tuned to predict 176 arXiv subject areas.  We compare the adapted score against three post‑hoc baselines: Maximum Softmax Probability (MSP), MaxLogit, and the temperature‑scaled free‑energy score originally proposed for vision models~\cite{EnergyOOD}.  Evaluation uses two deployment‑oriented metrics, Operational Open‑Set Accuracy (OOSA) and AUOSCR.

Our contributions are:
\begin{enumerate}
    \item \textbf{Adapting feature attenuation to NLP.} We present the first systematic transfer of COSTARR and the feature attenuation hypothesis to transformer‑based text classifiers.
    \item \textbf{Baseline study at scale.} We benchmark attenuation‑, logit‑, and energy‑based OSR scores on a 176‑class arXiv abstract task using BERT (base) and GPT-2.
    \item \textbf{Empirical findings.} We observe that COSTARR works in NLP without extra training but offers no clear advantage over MaxLogit or MSP, while the free‑energy score lags behind—suggesting model capacity rather than scoring choice may be the main limiting factor.
\end{enumerate}

\section{Related Works}

\subsection{Softmax and Logit-based OSR}

\emph{Maximum Softmax Probability (MSP)} is one of the most widely used baseline for OSR \cite{vaze2022opensetrecognitiongoodclosedset}. It works by thresholding the highest probability output by a standard softmax classifier. While intuitive and easy to implement, its effectiveness is limited by the tendency of neural networks to be overconfident on unfamiliar inputs, making it less reliable than more advanced OSR methods specifically designed to model uncertainty or the space of unknowns.

A simple way to mitigate some of the overconfidence issues of MSP is by using the maximum value among the raw logits (pre-softmax activations) output by the model for classification purposes. This method, called \emph{MaxLogit}, bypasses the need for a linear classification layer.

\subsection{Feature Attenuation in OSR}
Recent breakthroughs in computer vision reveal that classification layers suppress information critical for novelty detection—a phenomenon termed \emph{feature attenuation}.  The COSTARR framework \cite{COSTARR} formalizes this through the \emph{attenuation hypothesis}, which suggests that both pre‑attenuation features (deep embeddings before the final layer) and post‑attenuation features (after interaction with class weights) provide complementary signals for Open‑Set Recognition (OSR).  COSTARR leverages this by combining both signals into a unified scoring function that significantly outperforms prior OSR methods across image classification tasks using ViTs, ResNets, and ConvNeXts.  While effective in vision, its application to NLP has not been explored until now.

\subsection{Energy‑based OSR}
An alternative line of work swaps the usual soft‑max confidence for a \emph{free‑energy} score computed directly from the logits.  For an input $x$ with logit vector $f(x)=\bigl[f_1(x),\dots,f_K(x)\bigr]$ and a fixed temperature $T$, Liu \textit{et al.} \cite{EnergyOOD} define
\begin{equation*}
E(x;f)\;=\;-\,T\,\log\!\Bigl(\textstyle\sum_{k=1}^{K} e^{f_k(x)/T}\Bigr).
\label{eq:energy}
\end{equation*}
Because $-E(x;f)$ is a monotone transform of the soft‑max log‑partition function, higher values indicate more in‑distribution–like inputs.  At test time we simply threshold the \emph{negative} energy:
\[\text{score}(x)=-E(x;f),\qquad\text{reject if } \text{score}(x)<\tau,\]
with $\tau$ tuned on a small validation split containing known and unknown examples.  The method is completely post‑hoc and model‑agnostic, so we include it as a strong baseline in our transformer OSR experiments.

\subsection{Open‑Set Techniques for Text}
OSR for NLP remains underdeveloped relative to vision.  Leo and Kalita (2020) introduced one of the earliest open‑set approaches for text, focusing on incremental learning in authorship attribution.  Their CNN‑based pipeline emphasizes clustering and novelty detection but differs from our closed‑class training scenario.  Still, their work highlights the utility of outlier analysis and motivates evaluation strategies based on cluster separation and label discovery.

\subsection{OSR Evaluation Standards}
We adopt standardized OSR metrics to benchmark our method.  In particular, Operational Open‑Set Accuracy (OOSA) \cite{COSTARR} simulates deployment settings by determining decision thresholds on validation data that include both known and unknown samples.  Additionally, we report Area Under the Open Set Classification Rate Curve (AUOSCR) \cite{Agnostophobia}, which captures the trade‑off between recognizing known classes and rejecting unknowns across varying thresholds.  These metrics provide a rigorous framework to compare COSTARR against baseline methods such as softmax logit (MaxLogit), Maximum Softmax Probability (MSP), and the free‑energy score.

\section{Proposed Approach}

\begin{figure*}[!t]
    \centering
    \includegraphics[width=1\linewidth]{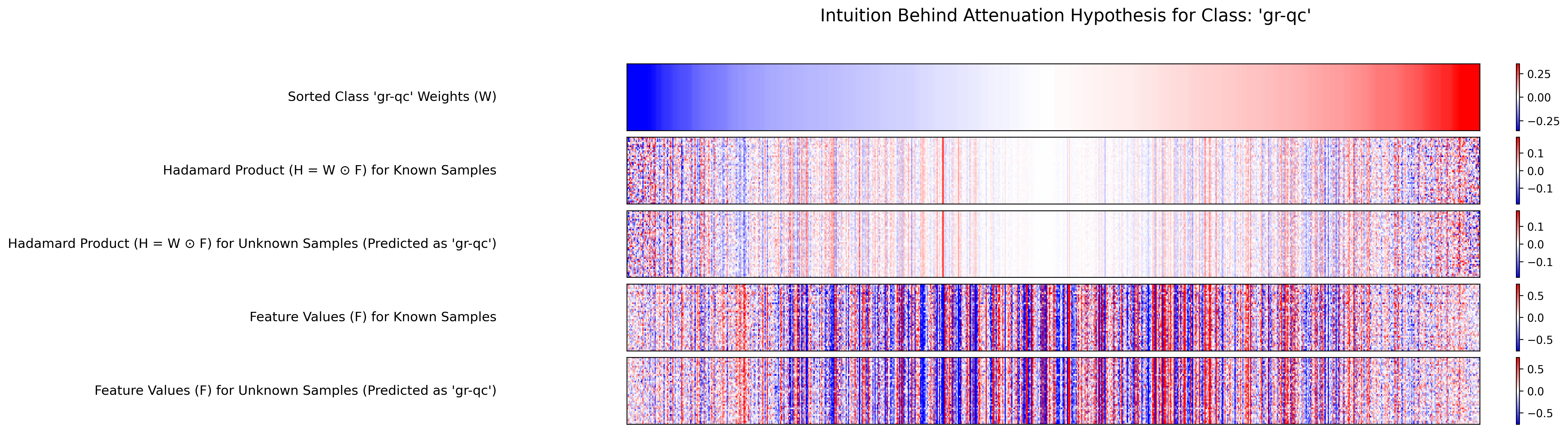}
    \caption{ Logit for class $j$ is formed from a dot product of its weights $W_j$ , and the input features $F$, but the low weighted features are attenuated and hence ignored by this class’s projection. The top color bar shows the weights $W$ for category 'gr-qc' from BERT (base)'s custom classification layer, sorted from low to high, left to right; the white zone denotes where weights approach zero. This sorting index orders all other color bars. The second bar shows the Hadamard Product $H$ for a known input $x_k$, and the 3rd row is for an unknown $x_u$, for which 'gr-qc' is the max logit. Each column represents a single feature dimension. The 4th and 5th bars show the feature vectors from the same test run and the most confident unknown samples from for that class, respectively. The consistency in low-saturation bars in the weights and Hadamard Product shows that regardless of what features are present, many dimensions are attenuated before classification}
    \label{weightsbert}
\end{figure*}

This section presents our adaptation of the COSTARR framework to transformer-based NLP models. We combine formal notation from the attenuation hypothesis with practical steps for implementation, training, and evaluation on the arXiv dataset.

\subsection{Initial Implementation Overview}

Our approach consists of four core steps:

\begin{enumerate}[label=(\arabic*)]
    \item \textbf{Dataset Selection}: Identify a labeled corpus of textual documents.
    \item \textbf{Model Training}: Fine-tune a transformer model with a custom linear classification head.
    \item \textbf{COSTARR Integration}: Implement COSTARR scoring using both pre- and post-attenuation features.
    \item \textbf{Evaluation}: Compare COSTARR to state-of-the-art OSR baselines such as MaxLogit, MSP, and free-energy score.
\end{enumerate}

\subsection{Dataset}

We use the Hugging Face \texttt{arxiv-abstracts-2021} dataset, which includes metadata for over 2 million preprints. These are categorized into 176 scientific domains called subcategories (ex: 'cs.AI' for artificial intelligence). Out of these, we randomly select 132 of them as known categories and treat the remaining 44 as unknown categories (a 0.75:0.25 ratio). Preprints belonging to no known category are considered out-of-distribution data for the purposes of OSR.

We randomly reserve $10\%$ of the documents across all categories for testing. The remaining documents from known categories are used for training.

\subsection{Model Selection and Feature Extraction}

We select two relatively lightweight pretrained models from Huggingface, BERT (base) and GPT-2, and added a custom classification head to enable COSTARR computation.

We extract deep features $F(x)$ from the penultimate layer ([CLS] embedding), and define post-attenuation features $H_j = F(x) \odot W_j$, where $W_j$ is the classification weight for class $j$. These features are used in the remaining steps to calculate the COSTARR score and perform evaluation.

\subsection{COSTARR Score for Transformers}

Given a test input $x$ and predicted class $m = \arg\max_j l_j(x)$:

\begin{align}
C_j(x) &:= \operatorname{Concat}(F(x), H_j) \\
\text{Sim}_j(x) &:= 0.5 \left(1 + \frac{C_j(x) \cdot \mu_{C_j}}{\|C_j(x)\| \cdot \|\mu_{C_j}\|} \right) \\
\lambda_m(x) &= \max_j \text{GNL}(l_j(x)) \\
\mathcal{S}(x) &= \lambda_m(x) \cdot \text{Sim}_m(x)
\end{align}

Here, $\mu_{C_j}$ is the class-wise mean of concatenated features computed from training data, and GNL is a global min-max normalization of logits.

\subsection{Evaluation Metrics}

We measure COSTARR's performance against commonly used methods for OSR such as MSP and free-energy score. MaxLogit is also included as a baseline for comparison, implemented through taking the maximum logit output by the LLM before the custom classification layer.

All methods are evaluated on the following metrics:

\begin{itemize}
    \item \textbf{OOSA (Operational Open-Set Accuracy)}: Threshold derived from a validation set (with held-out unknowns) to simulate deployment accuracy. We calculate OOSA at thresholds of 0, 0.1, 0.2, 0.3, $\ldots$, 1.0.
    \item \textbf{AUOSCR (Area Under OSCR Curve)}: Plots trade-off between known classification and unknown rejection over all thresholds.
\end{itemize}

\section{Results and Analysis}

For our experiments, we derive the optimal threshold for OOSA from a validation set of $10\%$ of the data for each model/score combination. From the test dataset, MaxLogit achieves an OOSA score of 0.319 for BERT (base) and 0.311 for GPT-2. On the other hand, COSTARR achieves an OOSA score of 0.319 for BERT (base) and 0.303 for GPT-2, while MSP achieves 0.320 for BERT (base) and 0.320 for GPT-2. Finally, free-energy achieves 0.172 for BERT (base) and 0.174 for GPT-2. Thus, there is no significant difference in OOSA scores between MaxLogit, COSTARR, and MSP, while free-energy lags substantially behind.

\begin{figure}[!h]
    \centering
    \includegraphics[width=1\linewidth]{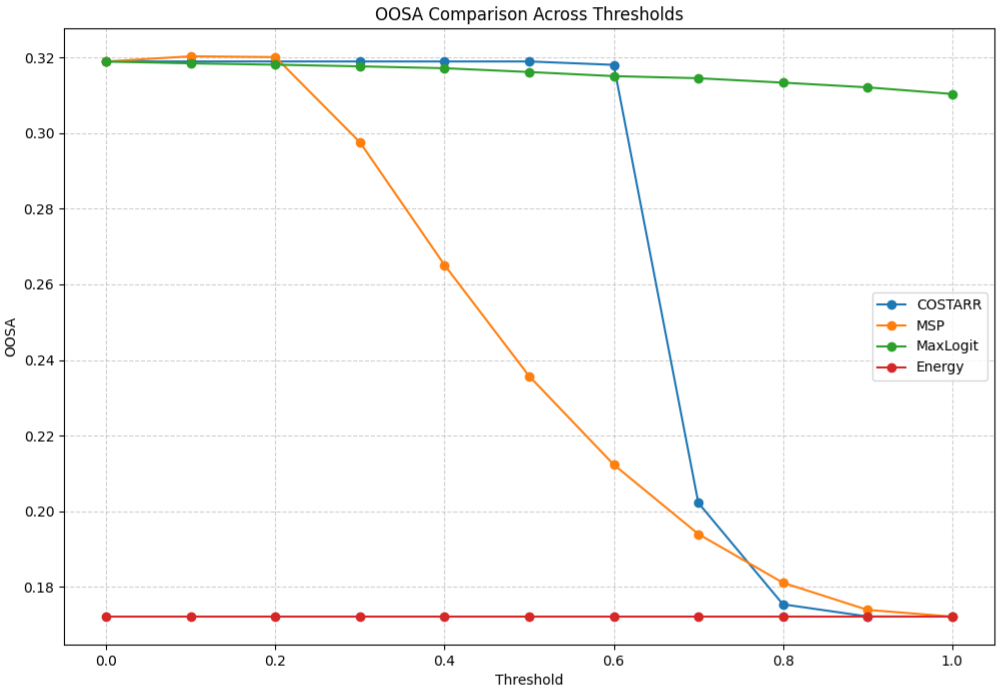}
    \caption{OOSA at various thresholds for BERT (base).}
    \label{OOSAbert}
\end{figure}

\begin{figure}[!h]
    \centering
    \includegraphics[width=1\linewidth]{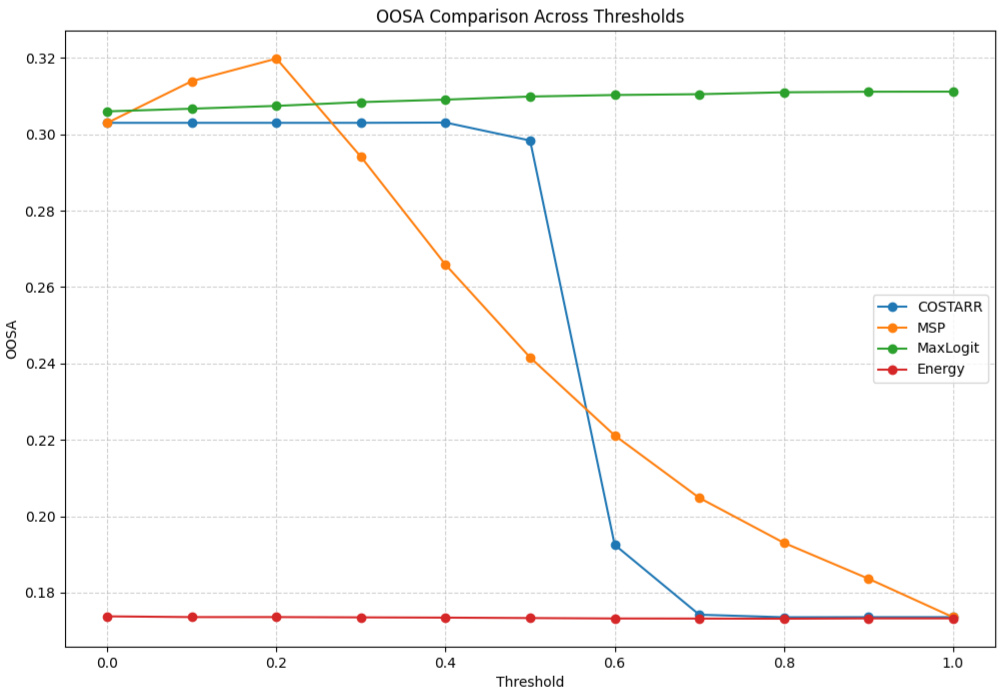}
    \caption{OOSA at various thresholds for GPT-2.}
    \label{OOSAgpt}
\end{figure}

However, interesting differences can be observed when we study the OOSA score at each threshold level (see figures \ref{OOSAbert} and \ref{OOSAgpt}). In both BERT (base) and GPT-2, we observe a substantial dropoff in MSP scores when the threshold exceeds 0.2, while COSTARR only substantially declines at thresholds above 0.5 or 0.6. This indicates that COSTARR is a more discriminatory approach that works for a larger range of thresholds compared to MSP. However, its performance is not significantly better than MaxLogit, whose OOSA score remains consistent across all thresholds.

\begin{figure}[!h]
    \centering
    \includegraphics[width=1\linewidth]{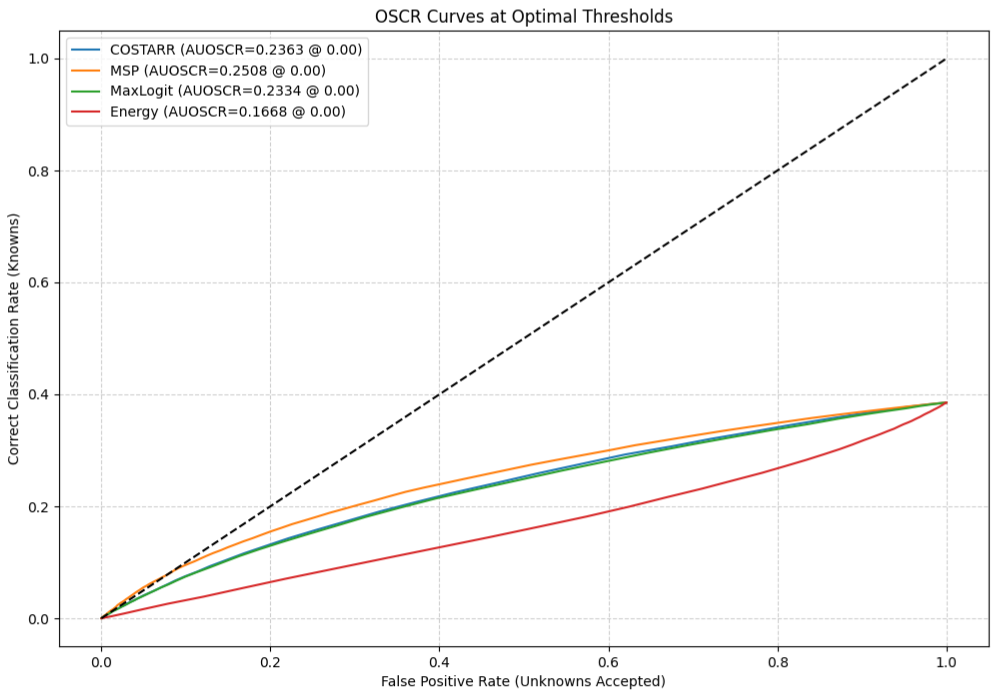}
    \caption{OSCR curves for BERT (base).}
    \label{AUOSCRbert}
\end{figure}

\begin{figure}[!h]
    \centering
    \includegraphics[width=1\linewidth]{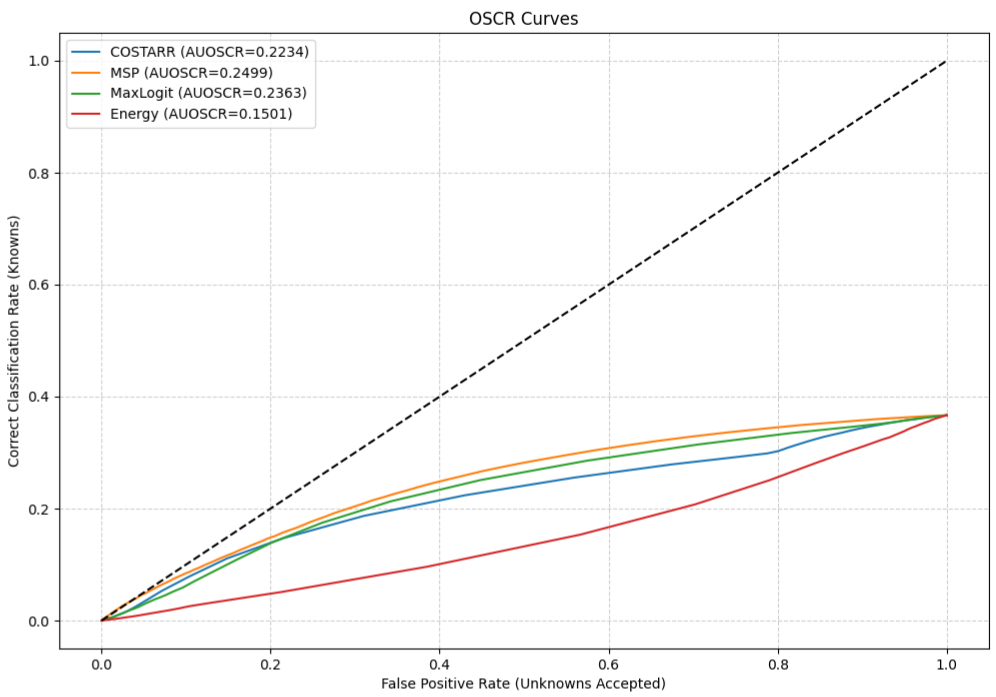}
    \caption{OSCR curves for GPT-2.}
    \label{AUOSCRgpt}
\end{figure}

We draw a similar conclusion when we look at AUOSCR scores. The OSCR curves for BERT (base) and GPT-2 are shown in figures \ref{AUOSCRbert} and \ref{AUOSCRgpt} respectively. COSTARR and MaxLogit achieved very similar AUOSCR scores. COSTAR has an AUOSCR of 0.236 for BERT (base) and 0.223 for GPT-2, while MaxLogit achieves 0.233 for BERT (base) and 0.236 for AUOSCR. On the other hand, MSP slightly outperformed the competition with an AUOSCR of 0.251 for BERT (base) and 0.250 for GPT-2. Finally, free-energy again performed poorly, netting an AUOSCR of only 0.167 for BERT (base) and 0.150 for GPT-2.

Despite these minor differences, it is worth noting that all of the various OSR approaches struggled with the task of arXiv category classification. We will address this observation in the next section. In the meantime, our results point towards the finding that none of the state-of-the-art OSR frameworks in classification settings generalize well to the NLP setting. At least when it comes to small, weak models like BERT and GPT-2 in complex settings with a large number of classes, they do not offer meaningful improvements over simply taking the MaxLogit before the classification layer.

\section{Limitations and Challenges}

It is important to recognize the limitations of our experiment. One of the challenges exposed by our findings is the poor performance of all the tested methodologies, including COSTARR. None of them achieve a OOSA or AUOSCR score above 0.4. Since all of the approaches struggled with the task, the issue likely lies not in the OSR methods, but in the LLM models themselves. BERT (base) and GPT-2 are both small and outdated models that may struggle to perform complex tasks such as arXiv category classification based on abstract. Certain abstracts from unknown categories like statistics may look very similar to abstracts from a known category like mathematics, and vice versa. This problem is exacerbated by the presence of 176 classes. It is worth noting that in preliminary experiments involving only 8 categories, we finetuned BERT (base) on a small subset of 200,000 samples and achieved an OOSA of 0.752 and an AUOSCR of 0.682. This demonstrates that the difficulty of the task drastically increases with larger numbers of classes. We would like to test our framework on larger and more powerful models such as GPT-4 and Deepseek-V3 to see if the performance improves.

Traditionally, OSR is a direction mostly focused in computer vision. One of our project's primary contributions is to bring this framework into the NLP setting. However, computer vision features may not directly translate to linguistic representations. The layer normalization and attention mechanisms of LLMs may suppress features differently.  Attenuation effects in transformers could differ fundamentally from CNNs.

Moreover, arXiv categories are unevenly distributed, and abstracts vary in length and structure. We may find it prudent to implement stratified sampling or abstract length thresholds. Furthermore, we would also like to test our findings on alternative datasets to validate whether our results are generalizable. Finally, NLP tasks are computationally intensive, and resource constraints can present great challenges especially as we scale up our project to include more models and more comprehensive ablation studies.

\section{Conclusion}
Our paper addresses a critical vulnerability in transformer-based text classifiers: their tendency toward overconfident misclassification of inputs from unknown categories under real-world open-set conditions. Deployed NLP systems (e.g., content moderation, academic search engines) frequently encounter novel categories. Our approach can help reduce hazardous overconfidence in such scenarios. Moreover, by revealing how classification layers suppress novelty-detection signals, we offer a pathway to design more transparent and adaptable transformers.

Through experimentation over classifying arXiv categories based on abstract, we observe no substantial improvement in the OOSA and AUOSCR metrics when using COSTARR and other state-of-the-art OSR methods. There appears to be no evidence that these powerful approaches in classification settings generalize well to NLP. For future research, we would like to test our framework on larger and more powerful LLMs such as GPT4 and Deepseek-V3 to see if these observations hold for high performing models. We also plan to conduct ablation studies to determine whether the feature attenuation hypothesis holds empirically in LLMs.

\bibliography{refs}

@InProceedings{MovingTowardsOpenSetIncrementalLearning,
author="Leo, Justin
and Kalita, Jugal",
editor="Arai, Kohei
and Kapoor, Supriya
and Bhatia, Rahul",
title="Moving Towards Open Set Incremental Learning: Readily Discovering New Authors",
booktitle="Advances in Information and Communication",
year="2020",
publisher="Springer International Publishing",
address="Cham",
pages="739--751",
isbn="978-3-030-39442-4"
}

@article{Awholisticviewofcontinuallearning...,
title = {A wholistic view of continual learning with deep neural networks: Forgotten lessons and the bridge to active and open world learning},
journal = {Neural Networks},
volume = {160},
pages = {306-336},
year = {2023},
issn = {0893-6080},
doi = {https://doi.org/10.1016/j.neunet.2023.01.014},
url = {https://www.sciencedirect.com/science/article/pii/S089360802300014X},
author = {Martin Mundt and Yongwon Hong and Iuliia Pliushch and Visvanathan Ramesh}
}

@INPROCEEDINGS {TowardsOpenSetDeepNetworks,
author = { Bendale, Abhijit and Boult, Terrance E. },
booktitle = { 2016 IEEE Conference on Computer Vision and Pattern Recognition (CVPR) },
title = {{ Towards Open Set Deep Networks }},
year = {2016},
volume = {},
ISSN = {1063-6919},
pages = {1563-1572},
doi = {10.1109/CVPR.2016.173},
url = {https://doi.ieeecomputersociety.org/10.1109/CVPR.2016.173},
publisher = {IEEE Computer Society},
address = {Los Alamitos, CA, USA},
month =Jun}

@article{ComparingBERT,
  author       = {Santiago Gonz{\'{a}}lez{-}Carvajal and
                  Eduardo C. Garrido{-}Merch{\'{a}}n},
  title        = {Comparing {BERT} against traditional machine learning text classification},
  journal      = {CoRR},
  volume       = {abs/2005.13012},
  year         = {2020},
  url          = {https://arxiv.org/abs/2005.13012},
  eprinttype    = {arXiv},
  eprint       = {2005.13012},
  bibsource    = {dblp computer science bibliography, https://dblp.org}
}

@unpublished{COSTARR,
  author  = {Anonymous},
  title   = {{COSTARR}: Consolidated Open Set Technique with Attenuation for Robust Recognition},
  note    = {Manuscript under review at International Conference on Computer Vision (ICCV)},
  year    = {2025}
}

@inproceedings{Agnostophobia,
 author = {Dhamija, Akshay Raj and G\"{u}nther, Manuel and Boult, Terrance},
 booktitle = {Advances in Neural Information Processing Systems},
 editor = {S. Bengio and H. Wallach and H. Larochelle and K. Grauman and N. Cesa-Bianchi and R. Garnett},
 pages = {},
 publisher = {Curran Associates, Inc.},
 title = {Reducing Network Agnostophobia},
 url = {https://proceedings.neurips.cc/paper_files/paper/2018/file/48db71587df6c7c442e5b76cc723169a-Paper.pdf},
 volume = {31},
 year = {2018}
}

@inproceedings{EnergyOOD,
  author       = {Weitang Liu and Xiaoyun Wang and John Owens and Yixuan Li},
  title        = {Energy-based Out-of-distribution Detection},
  booktitle    = {Advances in Neural Information Processing Systems},
  year         = {2020}
}

@misc{vaze2022opensetrecognitiongoodclosedset,
      title={Open-Set Recognition: a Good Closed-Set Classifier is All You Need?}, 
      author={Sagar Vaze and Kai Han and Andrea Vedaldi and Andrew Zisserman},
      year={2022},
      eprint={2110.06207},
      archivePrefix={arXiv},
      primaryClass={cs.CV},
      url={https://arxiv.org/abs/2110.06207}
}
\bibliographystyle{acl_natbib}

\end{document}